\documentclass[10pt,twocolumn,letterpaper]{article}

\usepackage{cvpr}
\usepackage{times}
\usepackage{epsfig}
\usepackage{graphicx}
\usepackage{amsmath}
\usepackage{amssymb}
\usepackage{booktabs}
\usepackage{multirow}
\usepackage{siunitx}
\usepackage{hyperref}

\usepackage{physics,amsmath}

\usepackage{booktabs}
\usepackage{enumitem}

\begin{document}

\title{Analyzing Bias in Diffusion-based Face Generation Models}
\author{Malsha V. Perera\\
Johns Hopkins University\\
{\tt\small jperera4@jhu.edu}
\and
Vishal M. Patel\\
Johns Hopkins University\\
{\tt\small vpatel36@jhu.edu}
}

\maketitle
\thispagestyle{empty}

\begin{abstract}
   Diffusion models are becoming increasingly popular in synthetic data generation and image editing applications. However, these models can amplify existing biases and propagate them to downstream applications. Therefore, it is crucial to understand the sources of bias in their outputs. In this paper, we investigate the presence of bias in diffusion-based face generation models with respect to attributes such as gender, race, and age. Moreover, we examine how dataset size affects the attribute composition and perceptual quality of both diffusion and Generative Adversarial Network (GAN) based face generation models across various attribute classes. Our findings suggest that diffusion models tend to worsen distribution bias in the training data for various attributes, which is heavily influenced by the size of the dataset. Conversely, GAN models trained on balanced datasets with a larger number of samples show less bias across different attributes. 
\end{abstract}

\section{Introduction}
Generative models have become increasingly popular in recent years and are now widely used in various real-world applications, ranging from image/video editing applications to synthetic data generation \cite{azizi2023synthetic, boutros2023synthetic}. These models can produce high-quality and realistic images, which has led to their increased adoption in various industries, including commercial art, entertainment and graphic design. While these models have shown impressive results in generating realistic images, they can also perpetuate and amplify existing biases which can have serious consequences for society. In particular, Generative Adversarial Networks (GANs) \cite{Goodfellow} have shown to exhibit bias \cite{maluleke2022ganracebias, JAIN2022103652}. Denoising Diffusion Probabilistic Models (DDPM) \cite{DDPM} is an emerging paradigm for generative modeling which learns the distribution of data through a Markovian process. Diffusion models have outperformed GANs in image synthesis and  have demonstrated a remarkable ability in generating high-quality images \cite{Diffusion_beat_GAN}. Despite their potential, diffusion models are likely to exhibit bias, especially when used for face generation tasks. Hence, it is crucial to investigate the circumstances under which diffusion models generate biased outputs and the root causes of such biases. Understanding these factors can help develop strategies to mitigate biases in diffusion models and promote fair and unbiased outcomes in various real-world applications.

In this work, we aim to understand bias in diffusion models in the context of the perceived attributes: gender, race and age. Our evaluation methodology does not rely on objectively determining the actual gender, race or age of the subject in the image, but rather on measuring the perceived attributes. The reason for this is that attributes such as gender and race are not solely determined by physical characteristics, but are also shaped by complex social factors. We conduct a comprehensive study to explore the impact of training dataset distribution on the imbalances in samples generated using diffusion models in terms of the considered attributes. Somepalli et al. \cite{somepalli2022diffusion} have raised concerns about the tendency of diffusion models to memorize the training data distribution, resulting in the generation of replicated training data. This phenomenon can have ethical implications, as popular diffusion models like Stable Diffusion \cite{Latentdiff} have been subject to intellectual property rights issues related to training data \cite{stable_diff_legal}. The replication of data is more pronounced in diffusion models trained on a limited number of training samples. Hence, it is important to investigate whether the number of training samples used to train DDPMs has any impact on bias. 

In this paper, we aim to examine the impact of training dataset size on potential imbalances in the samples generated by diffusion models that are trained on balanced datasets across different attributes.  Additionally, we have extended this study to GANs, in order to understand and compare the behaviour of both generative models in the above conditions. Moreover, we analyze whether generative models trained on balanced datasets exhibit differences in the perceptual quality of generated samples across different classes that belong to each attribute.


Based on our experiments, we have made several noteworthy observations. These include:
\begin{itemize}[noitemsep]
\item Diffusion-based face generative models amplify the bias in terms of the perceived gender, race, and age distribution in the training dataset.
\item When trained with balanced datasets of varying sizes, diffusion models demonstrate bias towards certain attribute classes, with the degree of bias being dependent on the specific dataset and its size.
\item  When using a balanced training dataset with a larger number of samples, GAN-based face generative models demonstrate a greater ability to preserve the attribute composition compared to diffusion models.
\end{itemize}

\section{Related Work}


\noindent\textbf{Generative Models.} 
Generative Adversarial Networks (GANs) \cite{Goodfellow} are a type of implicit generative model consisting of a generator and discriminator that are trained by optimizing a minimax objective. The discriminator's task is to distinguish between real and fake data, while the generator learns to produce fake data samples that are intended to be indistinguishable from real data samples. GANs have been used for various applications, including image and video generation, data augmentation, and image manipulation, due to their ability to produce high-quality images \cite{GAN_review_mingyu}.

Denoising Diffusion Probabilistic Models are a class of generative models that perform image synthesis through a $T$-step Markovian process \cite{DDPM}. DDPMs comprise of two processes - a forward process and a reverse process. The forward process sequentially transforms data to a white Gaussian noise by adding Gaussian noise in $T$ diffusion time steps. The reverse diffusion process starts from a standard Gaussian and denoises iteratively to generate a sample from the training distribution. Due to its remarkable ability in generating high quality images, diffusion models are widely used in image manipulation tasks \cite{SR3, Ho2022CascadedDM,Latentdiff, Diffusion_beat_GAN}.\\  
\noindent\textbf{Bias in Computer Vision}
has become a critical issue in recent years due to the widespread use of machine learning algorithms in various real-world applications. Numerous research efforts have aimed to evaluate, understand, and mitigate bias, particularly in face recognition-based applications. Numerous empirical studies have revealed that publicly available face recognition algorithms exhibit bias towards sensitive attributes, including gender and race \cite{pmlr-v81-buolamwini18a,fr_bias}. Albeiro \etal \cite{ijcb_albeiro_gender_fr} investigated the impact of gender distribution in training data on face recognition accuracy and found that gender-balanced training data does not necessarily lead to gender-balanced test accuracy. In \cite{S_2019_CVPR_Workshops}, Krishnapriya \etal analyzed the effects of race and skin tone on face recognition accuracy. Additionally, studies such as \cite{Albiero2021makeup,face_regions_fr,wu2023face} have explored face recognition performance under diverse factors such as skin brightness, hairstyles, and makeup. Several efforts have been undertaken to alleviate the biases that exist in face recognition algorithms. For instance, Wang \etal \cite{wang9156925} developed a framework to measure and mitigate intrinsic biases with respect to attributes such as gender in recognition tasks. Dhar \etal \cite{dhar} proposed to mitigate bias in face recognition with a distillation-based approach by enforcing the network to attend to similar face regions, regardless of the attribute category.\\ 
\noindent\textbf{Bias in GANs.}
Several studies have demonstrated that GANs are susceptible to bias. For instance, Jain \etal \cite{JAIN2022103652} conducted a study in which they found that widely used GAN-based face generation models can amplify biases related to gender and skin tone when trained on unbalanced datasets. Maluleke \etal \cite{maluleke2022ganracebias} conducted a comprehensive study to assess how the performance of GANs is affected by the racial composition of the training data. Their findings demonstrated that the racial composition of generated images successfully preserves that of the training data. Few recent studies have focused on developing methods to address the inherent bias in GANs. Yu \etal \cite{inclusivegan} interpret the problem of minority inclusion as a data coverage issue, and proposed Inclusive GAN to improve data coverage by harmonizing adversarial training with reconstructive generation. Another approach, proposed by Tan \etal \cite{tan2021improving}, involves retraining GAN models to generate  images with more balanced facial attributes. The proposed method involves shifting the interpretable semantic distribution in the latent space to achieve a more balanced image generation process. Kenfack \etal \cite{kenfack2022repfairgan} empirically attributed the unfairness in GANs to the difference in groups' gradient norms  during discriminator training and proposed a new training strategy based on gradient clipping in order to mitigate the bias.\\ 
\noindent\textbf{Bias in Diffusion Models.}
In recent years, several studies have focused on exploring potential biases in diffusion models, particularly in the context of text-to-image generation \cite{naik2023social, luccioni2023stable}. Luccioni \etal \cite{luccioni2023stable} proposed a new method for exploring and quantifying social
biases in text-to-image systems. The proposed method involves creating collections of generated images that highlight the variation of a text-to-image system across social attributes such as gender and ethnicity, as well as target attributes such as professions and gender-coded adjectives. Their analysis revealed that across target attributes, all the considered  text-to-image systems exhibited a substantial over-representation of the portion of their latent space that was associated with whiteness and masculinity. In a recent study, Naik \etal \cite{naik2023social}  analyzed the social biases that are commonly reflected in generated images by examining the depiction of occupations, personality traits, and everyday situations across representations of attributes such as gender, age, race, and geographical location. The authors discovered that text-to-image models exhibit significant biases across all of these attributes. 
To the best of our knowledge, no research has investigated the presence of bias in diffusion models concerning face generation while varying the size of the training dataset.\\
\noindent\textbf{Data Replication in Generative Models.}
Recently, several studies explored the issue of training data replication or memorization in generative models \cite{gan_replication, somepalli2022diffusion, carlini2023extracting}. In one such study, Feng \etal \cite{gan_replication} demonstrated that GANs have a tendency to replicate the training data, with the replication percentage decreasing exponentially in proportion to the complexity and size of the dataset. Somepalli \etal \cite{somepalli2022diffusion} utilized image retrieval frameworks to detect content replication in images generated using diffusion models and analyzed how various factors, including training set size, affect the extent of content replication. A similar study conducted by Carlini \etal \cite{carlini2023extracting}  showed that diffusion models memorize and replicate training data. Furthermore, their study demonstrated that diffusion models are much less private and leak more training data than generative models such as GANs. 

\section{Method}


 In this section, we provide an overview of the training datasets used, their attribute distributions, and the specific subsets of training data employed to study biases in generative models. Our analysis focuses on biases related to gender, race, and age attributes in two widely popular datasets: FFHQ \cite{FFHQ} and FairFace \cite{fairface}. We examine the existence of bias in diffusion models by utilizing a lighter version of ADM introduced in \cite{p2} as the network architecture for unconditional face generation. Additionally, we analyze the presence of bias in GAN-based face generation by using Style-GAN2-ADA \cite{stylegan}.\\

\begin{figure}
    \centering
    \includegraphics[width=.8\linewidth]{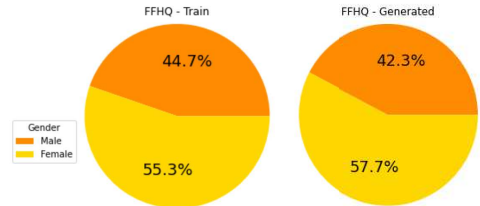}
    \vskip-16pt\caption{\textbf{Gender distribution in the FFHQ} training data (left) and generated data (right) using diffusion model.}
    \label{FFHQ_train_gender}
    \vspace{-5pt}
\end{figure}

\begin{figure}
    \centering
    \includegraphics[width=.8\linewidth]{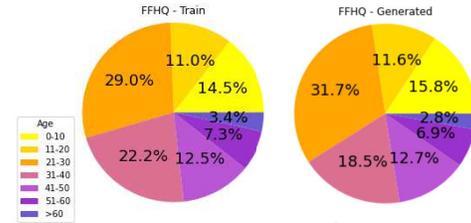}
    \vskip-10pt\caption{\textbf{Age distribution in the FFHQ} training data (left) and generated data (right) using diffusion model.}
    \label{FFHQ_train_age}
\end{figure}

\noindent The Flickr Faces HQ {\bf{(FFHQ Dataset)}} is a widely used dataset for face generation tasks \cite{FFHQ}. This dataset contains 70,000 aligned and cropped high-quality $1024\times 1024$ resolution images scraped from Flickr. We use the FFHQ dataset to evaluate the presence of bias in face generation models across gender and age attributes. Out of 70000 face images in the FFHQ dataset,  $55.3\%$ and $44.7\%$ of the images are categorized as females and males respectively \footnote{https://github.com/DCGM/ffhq-features-dataset}. The FFHQ dataset is composed of $14.5\%, 11\%, 29\%, 22.2\%, 12.6\%, 7.2\%,$ and $3.4\%$ samples in the age ranges 0-10,11-20,21-30,31-40,41-50,51-60 and above 60, respectively. Figure \ref{FFHQ_train_gender} and \ref{FFHQ_train_age} illustrates the gender, and age distribution for the FFHQ dataset.

\begin{figure}
    \centering
    \includegraphics[width=.8\linewidth]{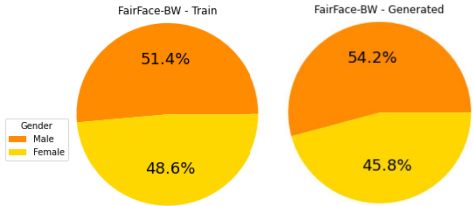}
   \vskip-15pt \caption{\textbf{Gender distribution in the FairFace-BW} training data (left) and generated data (right) using diffusion model.}
    \label{Fairface_train_gender}
\end{figure}

\begin{figure}
    \centering
    \includegraphics[width=.8\linewidth]{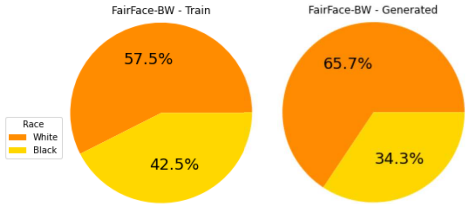}
    \vskip-15pt\caption{\textbf{Racial distribution in FairFace-BW} training data (left) and generated data (right) using diffusion model.}
    \label{Fairface_train_race}
    \vspace{-10pt}
\end{figure}

\noindent The \textbf{FairFace dataset}, which is widely used for studying bias in computer vision, consists of 108,501 facial images that have been aligned and cropped, covering seven racial/ethnic groups and two genders \cite{fairface}. The dataset contains images of non-public figures, reducing selection bias, and aims to enhance fairness and accuracy in face recognition systems by offering a more diverse and inclusive dataset. For our investigation, we focus on two racial categories in the FairFace dataset, labeled as `Black' and `White', and refer to this subset of data as FairFace-BW. While we acknowledge the importance of studying bias across all racial categories, our study is a preliminary investigation, and we have chosen to simplify the categorization in this manner. We utilize the FairFace-BW dataset, which contains $28,760$ face images to examine the racial and gender biases in face generation. Of these, $16,527$ ($57.47\%$) are categorized as White, while $12,233$ ($42.53\%$) are labeled as Black. Moreover, the Fairface-BW dataset includes $51.4\%$ male and $48.6\%$ female face images. Figure \ref{Fairface_train_race} and \ref{Fairface_train_gender} show the racial, and gender distribution for the FairFace-BW dataset.

We use the complete FFHQ dataset to train a diffusion model for generating faces and analyze the age and gender distribution of the generated face images.  Similarly, we train a diffusion-based face generation model using the FairFace-BW dataset and quantify the racial and gender distribution of the generations. To understand the effect of dataset size on the bias present in face generation models, we train multiple diffusion and GAN models using subsets of the FFHQ and FairFace-BW datasets. The subsets of training data are randomly sampled from the respective datasets to create balanced datasets with respect to the attribute being considered. We employ several attribute classifier networks that are fine-tuned on the FFHQ and FairFace-BW datasets to determine the perceived distribution of attributes.

\section{Experiments and Results}

\begin{figure*}
    \centering
    \includegraphics[width=0.9\linewidth]{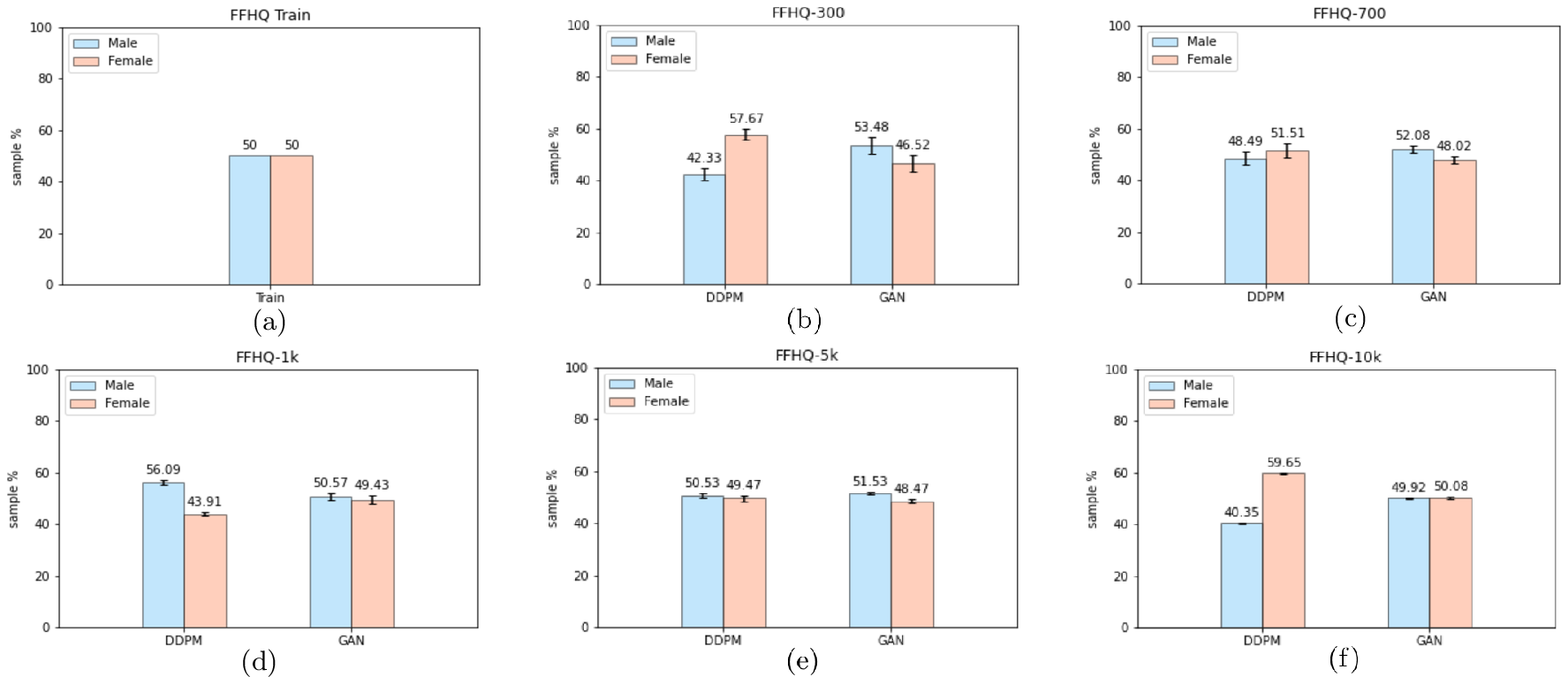}
   \vskip-4pt \caption{\textbf{Gender Distribution of FFHQ training data subsets and generated data :} a) gender distribution in training subset, results on GAN and diffusion models trained with b) 300 samples, c) 700 samples, d) 1000 samples, e) 5000 samples, f) 10000 samples}
    \label{gender_FFHQ_datasize}
\end{figure*}

\begin{figure*}
    \centering
    \includegraphics[width=0.9\linewidth]{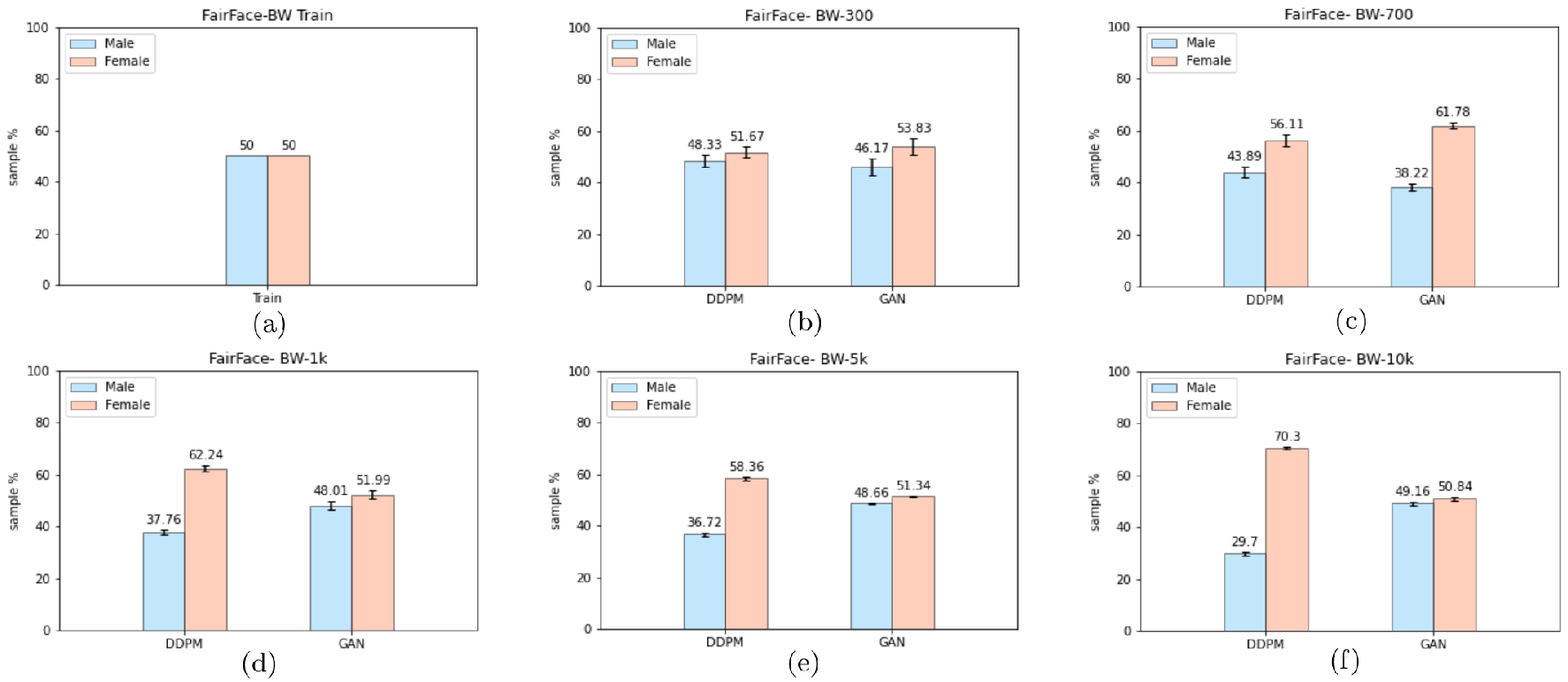}
    \vskip-5pt\caption{\textbf{Gender Distribution of FairFace-BW training data subsets and generated data :} a) gender distribution in training subset, results on GAN and diffusion models trained with b) 300 samples, c) 700 samples, d) 1000 samples, e) 5000 samples, f) 10000 samples}
    \label{gender_fairface_datasize}
\end{figure*}

\begin{figure*}
    \centering
    \includegraphics[width=.8\linewidth]{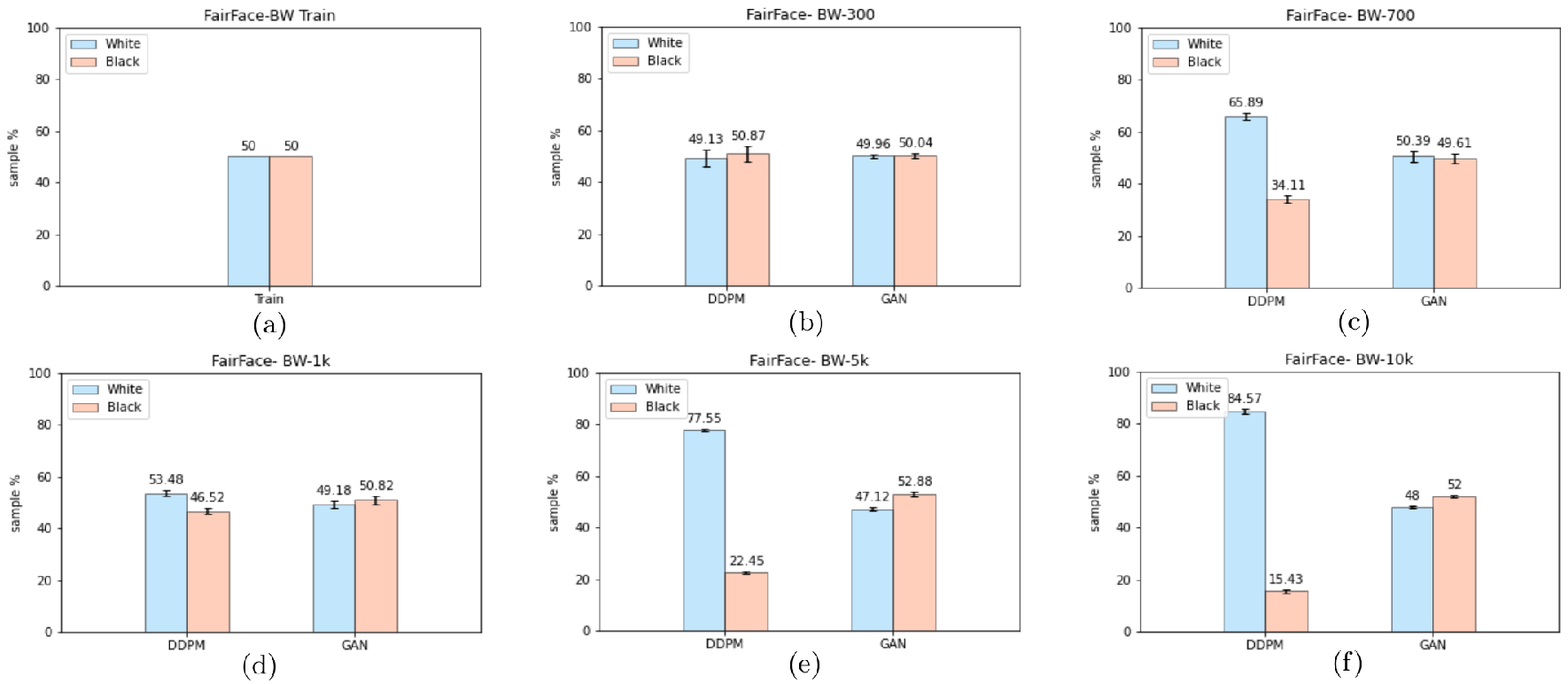}
   \vskip-5pt \caption{\textbf{Racial Distribution of FairFace-BW training data subsets and generated data :} a) racial distribution in training subset, results on GAN and diffusion models trained with b) 300 samples, c) 700 samples, d) 1000 samples, e) 5000 samples, f) 10000 samples}
    \label{race_fairface_datasize}
\end{figure*}

\begin{figure*}
    \centering
    \includegraphics[width=.9\linewidth]{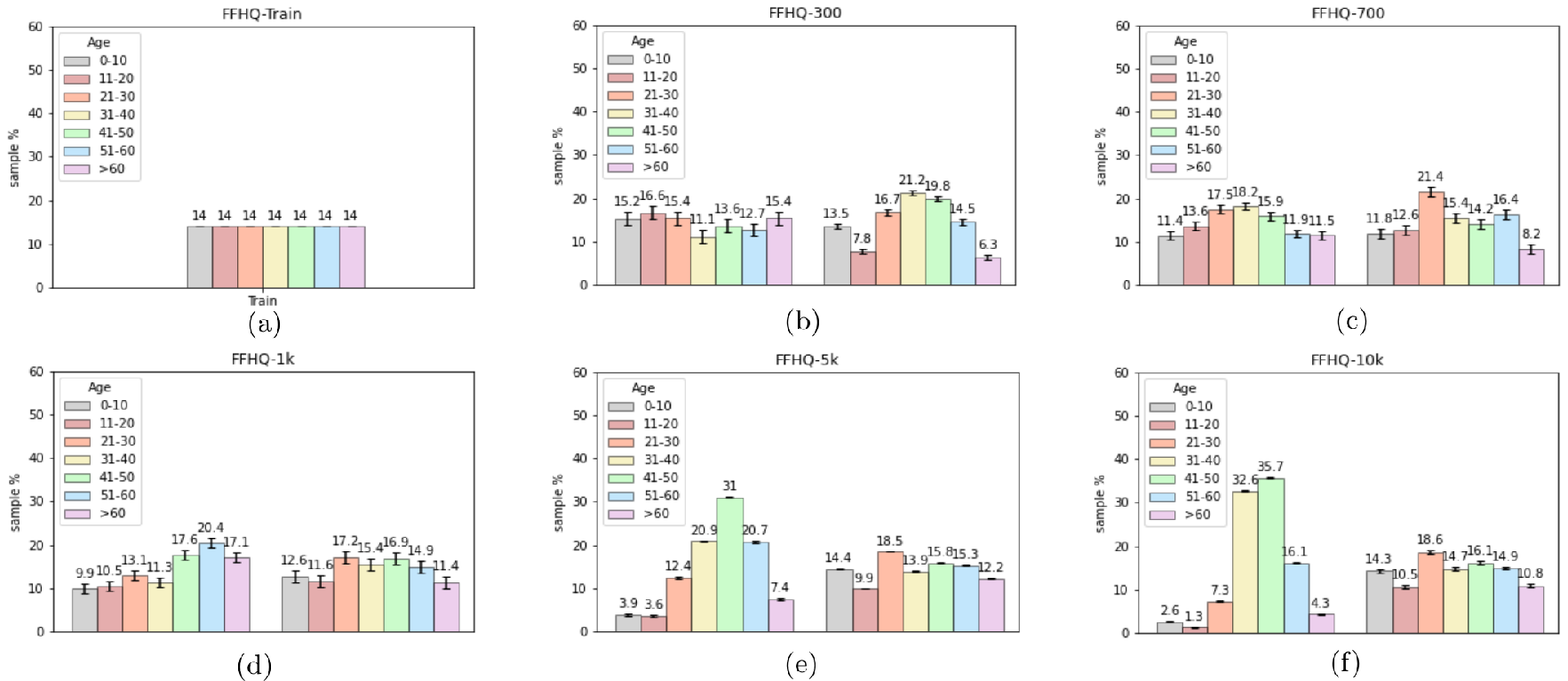}
   \vskip-5pt \caption{\textbf{Age Distribution of FFHQ training data subsets and generated data :} a) age distribution in training subset, results on GAN and diffusion models trained with b) 300 samples, c) 700 samples, d) 1000 samples, e) 5000 samples, f) 10000 samples}
    \label{age_FFHQ_datasize}
\end{figure*}

To ensure high-quality generations, we train all the diffusion models for a minimum of 1M steps. To quantify the attribute distribution, we then generate the same number of samples as the number of samples used to train the model. All models are trained on a single NVIDIA RTX A5000 GPU using images of size $256\times256$.

\subsection{Attribute Distribution of Diffusion-based Face Generations}
We analyze the perceived gender and age distribution of face images generated by the diffusion model trained on the complete FFFHQ dataset. Fig. \ref{FFHQ_train_gender} shows that 58\% of the generated samples are perceived as females and 42\% are perceived as males.  As per Figure \ref{FFHQ_train_age}, the generated samples are distributed across the age ranges as follows: 14.5\% in the range of 0-10, 11\% in the range of 11-20, 29\% in the range of 21-30, 22.2\% in the range of 31-40, 12.6\% in the range of 41-50, 7.2\% in the range of 51-60, and 3.4\% for ages above 60.

Furthermore, we determine the perceived distribution of gender and race in the face images generated by the diffusion model trained on the FairFace-BW dataset. Analysis presented in Figure \ref{Fairface_train_race} reveals that 66\% of the generated images are perceived as White while 34\% of the generated images are perceived as Black. Figure \ref{Fairface_train_gender} shows that the gender attribute of the generated samples from this diffusion model is distributed as 54\% male and 46\% as female.

\subsection{Effect of Dataset Size on Attribute Bias in Face Generation Models}
We investigate the impact of dataset size on attribute bias in diffusion and GAN-based face generation models. To achieve this, we train diffusion and GAN models on training subsets of the respective dataset, containing 300, 700, 1000, 5000, and 10000 samples for each selected attribute. These training subsets are created by random sampling from the corresponding dataset to achieve a balanced distribution of the considered attribute. We then generate an equal number of samples as the training subset, repeating this process five times to obtain the mean attribute distribution of the generated data.

The training data subsets used to investigate the impact of dataset size on gender bias in face generation models were sampled to have an equal distribution of 50\% males and 50\% females. Figure \ref{gender_FFHQ_datasize} and Figure \ref{gender_fairface_datasize} show the gender distributions of face images generated by diffusion and GAN-based models trained on different-sized subsets of the FFHQ and FairFace-BW datasets, respectively. 

To examine the impact of dataset size on racial bias in face generation models, we utilize the FairFace-BW dataset. We randomly sample images from this dataset to create training data subsets that have a balanced racial composition of 50\% Black and 50\% White. Figure \ref{race_fairface_datasize} displays the racial distribution of the generated face images using diffusion and GAN-based models at different sizes of the FairFace-BW training subsets.

We analyze the influence of dataset size on age distribution in face generation models based on diffusion and GAN architectures. To this end, we train these models on subsets of the FFHQ dataset with varying numbers of samples. Each subset is composed of face images equally distributed among the age ranges, 0-10, 11-20, 21-30, 31-40, 41-50, 51-60, and above 60. In other words, each age range is roughly 14.3\% of the training subset. Figure \ref{age_FFHQ_datasize} illustrates the age distribution of the diffusion and GAN-based face generations for the training subset sizes 300, 700, 1000, 5000, and 10000. 

\subsection{Face Generation Quality}
The Frechet inception distance (FID) \cite{FID} is a widely used metric for evaluating the perceptual quality of images generated by a generative model. FID measures the quality of images by comparing distributions of features extracted using the Inception network between training and generated images. Lower FID scores indicate higher perceptual quality of the images. In this study, we examine the impact of dataset size on the perceptual quality of diffusion and GAN-based face generation models across different attribute classes using the FID score. We report the FID scores of generated samples across different attribute classes related to gender, race, and age in Tables \ref{gender_table}, \ref{race_table}, and \ref{age_table}, respectively.

\begin{table}[h]
\setlength\tabcolsep{0pt}
\vspace{-1pt}
\centering
\smallskip
\begin{tabular*}{\columnwidth}{@{\extracolsep{\fill}}lSSSS}
    \toprule
    \multirow{2}{*}{} &
      \multicolumn{2}{c}{Male} &
      \multicolumn{2}{c}{Female} \\
      & {DDPM} & {GAN} & {DDPM} & {GAN}\\
      \midrule
    FFHQ-300 & \text{72.73} & \text{105.27}  & \text{66.28} & \text{114.11}  \\
    FFHQ-700 & \text{76.72} & \text{76.56}  & \text{75.38} & \text{84.89} \\
    FFHQ-1000 & \text{83.12} & \text{61.79}  & \text{78.78} & \text{65.72} \\
    FFHQ-5000 & \text{64.12} & \text{25.72}  & \text{66.33} & \text{25.64} \\
    FFHQ-10000 & \text{39.31} & \text{17.85}  & \text{44.81} & \text{18.68} \\
    \midrule
    FairFace-BW-300 & \text{55.23} & \text{124.91}  & \text{60.32} & \text{115.52}  \\
    FairFace-BW-700 & \text{65.03} & \text{95.57}  & \text{70.96} & \text{76.71} \\
    FairFace-BW-1000 & \text{64.71} & \text{69.46}  & \text{57.97} & \text{63.04} \\
    FairFace-BW-5000 & \text{88.09} & \text{26.60}  & \text{80.78} & \text{26.88} \\
    FairFace-BW-10000 & \text{48.07} & \text{19.34}  & \text{49.11} & \text{19.17} \\
    \bottomrule
  \end{tabular*}
  \vspace{1mm}
  \caption{FID scores of the images generated by models trained on gender-balanced training data at different dataset sizes. }
  \label{gender_table}
  \vspace{-3pt}
\end{table}

\begin{table}[h]
\setlength\tabcolsep{0pt}
\vspace{-1pt}
\centering
\smallskip
\begin{tabular*}{\columnwidth}{@{\extracolsep{\fill}}lSSSS}
    \toprule
    \multirow{2}{*}{} &
      \multicolumn{2}{c}{White} &
      \multicolumn{2}{c}{Black} \\
      & {DDPM} & {GAN} & {DDPM} & {GAN}\\
      \midrule
    FairFace-BW-300 & \text{62.86} & \text{112.81}  & \text{70.54} & \text{121.99}  \\
    FairFace-BW-700 & \text{68.26} & \text{79.66}  & \text{80.20} & \text{79.31} \\
    FairFace-BW-1000 & \text{60.05} & \text{69.14}  & \text{68.38} & \text{71.99} \\
    FairFace-BW-5000 & \text{90.91} & \text{28.31}  & \text{105.87} & \text{25.69} \\
    FairFace-BW-10000 & \text{96.63} & \text{18.04}  & \text{88.72} & \text{17.02} \\
    \bottomrule
  \end{tabular*}
  \vspace{1pt}
  \caption{FID scores of the images generated by models trained on race-balanced training data at different dataset sizes}
  \label{race_table}
  \vspace{-10pt}
\end{table}

\begin{table*}[h]
\setlength\tabcolsep{0pt}
\vspace{-1pt}
\centering
\smallskip
\begin{tabular*}{\linewidth}{@{\extracolsep{\fill}}lSSSSSSSSSSSSSS}
    \toprule
    \multirow{2}{*}{} &
      \multicolumn{2}{c}{Age 0-10} &
      \multicolumn{2}{c}{Age 11-20} &
      \multicolumn{2}{c}{Age 21-30} &
      \multicolumn{2}{c}{Age 31-40}&
      \multicolumn{2}{c}{Age 41-50}&
      \multicolumn{2}{c}{Age 51-60} &
      \multicolumn{2}{c}{Age above 60}\\
      & {DDPM} & {GAN} & {DDPM} & {GAN} & {DDPM} & {GAN} & {DDPM} & {GAN} & {DDPM} & {GAN} & {DDPM} & {GAN} & {DDPM} & {GAN}\\
      \midrule
    FFHQ-300 & \text{106.5} & \text{185.30}  & \text{109.0} & \text{172.62} & \text{92.0} & \text{180.28} & \text{92.4} & \text{149.43} & \text{97.0} & \text{143.63} & \text{77.37} & \text{139.97} & \text{63.42} & \text{164.87}  \\
    FFHQ-700 & \text{112.45} & \text{137.34}  & \text{111.13} & \text{127.73} & \text{101.65} & \text{122.23} & \text{104.84} & \text{115.60} & \text{109.03} & \text{114.00} & \text{93.76} & \text{97.76} & \text{100.67} & \text{128.76}  \\
    FFHQ-1000 & \text{122.12} & \text{119.60}  & \text{125.32} & \text{123.50} & \text{134.36} & \text{112.37} & \text{132.68} & \text{107.99} & \text{116.44} & \text{103.31} & \text{95.24} & \text{91.53} & \text{107.99} & \text{102.41}  \\
    FFHQ-5000 & \text{185.75} & \text{58.81}  & \text{165.25} & \text{62.30} & \text{137.41} & \text{52.03} & \text{133.37} & \text{53.80} & \text{137.57} & \text{48.15} & \text{145.15} & \text{45.33} & \text{145.61} & \text{47.86}  \\
    FFHQ-10000 & \text{142.08} & \text{39.84}  & \text{147.70} & \text{43.79} & \text{118.69} & \text{37.37} & \text{130.81} & \text{37.01} & \text{124.94} & \text{33.31} & \text{123.48} & \text{30.47} & \text{133.50} & \text{35.44}  \\
    \bottomrule
  \end{tabular*}
  \vspace{1pt}
  \caption{FID scores of the images generated by models trained on age-balanced training data at different dataset sizes.}
  \label{age_table}
  \vspace{-8pt}
\end{table*}

\subsection{Attribute Classifiers}
In this study, we employed attribute classifiers based on ResNet-18 architecture \cite{ResNet18} trained on the corresponding datasets (FFHQ or FairFace) to evaluate the attribute distribution of the generated face images. In order to classify attributes in the generative models trained on the complete FFHQ dataset or its subsets, we trained two separate classifier networks for gender and age, using 69700 images from the FFHQ dataset. We allocated the remaining 300 images for testing the two classifiers. The FFHQ-based gender and age classification networks achieved an accuracy of 99.9\% and 95\% respectively.   


Furthermore, we also trained classifiers to determine attribute classes of the images generated using diffusion and GAN models trained on the FairFace-BW dataset or its subsets. The FairFace-BW-based classifiers were evaluated on images belonging to the Black and White classes of the validation set in the FairFace dataset. The FairFace-based gender and race classification networks achieved an accuracy of 99.15\% and 97.0\% respectively.


\section{Discussion}
As illustrated in Figure \ref{FFHQ_train_gender}, the diffusion model trained on the complete FFHQ dataset generates a greater percentage of female images compared to the gender distribution in the training data, indicating a bias towards female face image generation and potentially exacerbating training data bias. However, as shown in Figure \ref{Fairface_train_gender}, when we employ the diffusion model trained on the FairFace-BW dataset to generate face images, it demonstrates a bias towards male images. This highlights that the direction of bias towards a specific gender class is dependent on the data distribution in the training dataset.

In Figure \ref{FFHQ_train_age}, we observe the age distribution of the generated data produced by the diffusion model trained on the complete FFHQ dataset. While most age groups have a similar sample percentage as the training data, the proportion of faces in age groups above 50 has decreased compared to the training distribution. This suggests that the diffusion model trained on FFHQ is biased towards generating younger face images.

The results depicted in Figure \ref{Fairface_train_race} indicate that the diffusion model trained on the FairFace-BW dataset generates a higher proportion of face images belonging to the White racial class, suggesting that the model may perpetuate bias towards this particular group.


In order to investigate the impact of dataset size on attribute bias in face generation models, we trained all diffusion and GAN-based models using balanced datasets. This approach was taken to mitigate the potential impact of training data distribution bias on our analysis of the effect of dataset size on model bias. Figure \ref{gender_FFHQ_datasize} reveals that diffusion models trained on subsets of the FFHQ dataset exhibit the highest gender bias when trained with the dataset sizes of 300 and 10000. In these cases, the generated face images are predominantly female. In contrast, diffusion models trained on 700 and 5000 sized subsets of the FFHQ data generate gender distributions that are more comparable to the training data distribution. Notably, the diffusion model trained with 1000 samples from the FFHQ generates more male images than female images. Unlike the diffusion models trained on the FFHQ subsets, the generated gender distribution of diffusion models trained on the FairFace-BW subsets presented in Figure \ref{gender_fairface_datasize} is more biased towards females and this bias is amplified as the training subset size increases.


The gender distributions of GAN-based face generation models trained on the FFHQ or FairFace-BW subsets larger than 1000 samples are similar to the balanced training distribution, unlike diffusion models. However, when trained on lower sample sizes, specifically 300 and 700, GAN models exhibit more pronounced gender bias. It's worth noting that the bias varies depending on the dataset used; for example, GAN models trained on FFHQ generate more male images, while the bias towards generating female images is more pronounced in the case of FairFace-BW.


The results in Figure \ref{race_fairface_datasize} show that diffusion models trained on subsets of the FairFace-BW dataset exhibit a strong bias towards the White racial class, with extreme bias observed for subset sizes of 5000 and 10000 samples. However, GAN-based models trained on FairFace-BW consistently generate a racial distribution similar to the balanced training data distribution, regardless of subset size. This finding is consistent with a previous study by Maluleke et al. \cite{maluleke2022ganracebias} which suggests that GANs can preserve the racial composition of the FairFace dataset.


Based on Figure \ref{age_FFHQ_datasize}, the age attribute distribution in diffusion models trained with subsets of the FFHQ dataset occasionally exhibits consistency with the training distribution across different age groups. However, there is no apparent pattern of age distribution that remains consistent across the diffusion models trained with various data subset sizes. In contrast, GAN-based face generation models trained on different dataset sizes generally align with the training distribution for most age groups, except for the age groups of 11-20 and above 60, which are consistently generated in lower proportions regardless of dataset size.


In general, the diffusion models showed greater variation in attribute distribution compared to the balanced distribution in the training data, and this bias was dependent on the dataset and its size. In contrast, GAN-based models showed minimal bias, especially for larger dataset sizes, suggesting that they are better at preserving the attribute composition of the training data. The complexity of the chosen training subset was found to play a significant role in the variation of attribute distribution in diffusion models. Other factors such as lighting conditions, hair, and makeup also contribute to the complexity of the data and can make it more difficult for diffusion models to learn a balanced representation of the data. However, GANs showed significantly less bias in the generated images compared to diffusion models, indicating their robustness to these types of variations in the training data.

\begin{figure}[htp!]
    \centering
    \includegraphics[width=0.9\linewidth]{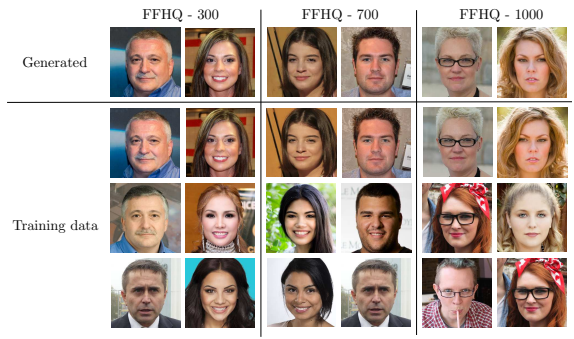}
    \caption{\textbf{Data replication in diffusion models}: row 2 to 4 present the top-3 matches from the training data.}
    \label{data_replication}
    \vspace{-5mm}
\end{figure}

Several studies in the literature have investigated the phenomenon of data replication in diffusion-based generative models. These studies have demonstrated that data replication is more common in models trained with less data \cite{gan_replication, somepalli2022diffusion}. One possible explanation for this phenomenon is that the model complexity exceeds the complexity of the training data, resulting in overfitting to the training data. Figure \ref{data_replication} illustrates examples of replicated samples produced by diffusion models trained with subsets of the FFHQ and FairFace-BW datasets. Consistent with previous studies, we observe that more replications occur in diffusion models trained with smaller numbers of samples. Our experiments reveal that data replication in diffusion models does not necessarily result in the replication of the attribute distribution of the training data. Rather, certain samples are replicated more frequently, leading to bias. Therefore, data replication in diffusion models trained with smaller sample sizes can affect bias with respect to different attributes. Although the study conducted by Feng \etal \cite{gan_replication} suggests that GANs also tend to replicate training data when trained on smaller dataset sizes, we observe that data replication is not as pronounced in GANs as it is in diffusion models. 

Table \ref{gender_table} illustrates that diffusion models trained on varied sizes of training subsets show bias across male and female classes, as the perceptual quality across the two classes differs. However, the class with the lowest FID score depends on the dataset and its size. For GAN-based models trained on FFHQ subsets, the perceptual quality of male images is either higher or comparable to that of female images. On the other hand, for GAN models trained on FairFace-BW data subsets, the female perceptual quality is higher or comparable to the perceptual quality of the male images. As shown in Table \ref{race_table}, diffusion-based face generative models, except for the model trained with 10000 samples, generate images of the white racial class with higher perceptual quality. However, the perceptual quality of the black and white classes for GAN models trained on FairFace-BW subsets depends on the dataset size. Table \ref{age_table} presents a comparison of FID scores for age groups in diffusion and GAN-based face generation models trained on different dataset sizes. The results indicate that in diffusion models trained on FFHQ subsets, the FID scores for generated images belonging to age groups above 21 are generally lower compared to those belonging to age groups under 20. In GAN-based face generation models, images belonging to age groups above 60 and below 20 tend to have lower perceptual quality, as reflected by their higher FID scores.

Overall, we can see that the FID scores for attribute classes in images produced by diffusion models have a pattern of starting low, then increasing, and finally decreasing as the size of the dataset used for training increases.The low FID scores in diffusion models trained with the lowest number of samples might be due to data replication, which causes the training data and generated data distribution to be more similar. In contrast, results from the GAN-based models show a gradual decrease in the FID score as the dataset size increases.

In our study, we employed automated classifiers to avoid the tedious task of manually determining attribute classes. While we recognize that these classifiers may exhibit bias towards certain attributes, we found that they showed an accuracy above 95\% when tested on a subset of the datasets we considered, and the classification results were not biased towards a particular category within a given attribute. (This can be seen from the confusion matrices presented in the Supplementary materials). Since we trained these classifier networks using the same datasets as the generative models, it is reasonable to assume that the classifiers exhibit minimal bias when determining the attribute class of the generated samples. Furthermore, it can be assumed that if there is any bias present, it will be minimal for generative models trained with less data that replicate the training data.

Our study highlights that diffusion-based face generation models have the potential to amplify bias present in training data, especially concerning gender, race, and age attributes. Therefore, researchers should be cautious while utilizing diffusion-based face generation models in downstream tasks, particularly those that require balanced models. Furthermore, dataset size and complexity should be taken into account while employing diffusion models for face generation since these models exhibit bias even when trained on balanced data. This emphasizes the need for bias mitigation techniques for diffusion-based face generation methods. Future research can include exploring bias in different diffusion-based models such as Latent Diffusion Models (LDM) \cite{Latentdiff} and studying the impact of other attributes such as hair and makeup. Additionally, examining how other generative models such as Variational Autoencoders (VAE) \cite{vae} perform on this task could be an interesting direction for future research.

\section{Conclusion}

Understanding and addressing the biases in diffusion-based face generation models is essential to mitigate the potential negative impact of these models in various applications. In this study we investigated the bias exhibited by diffusion-based face generation models across the attributes: gender, race and age. Furthermore we studied the impact of dataset size on the attribute composition and perceptual quality of diffusion and GAN-based face generation models across various attribute classes. Our results indicate:
\begin{itemize}[noitemsep]
\item diffusion models exacerbate the distribution bias in the training data in various attributes across different datasets.
\item bias in  the attribute distribution of diffusion models is heavily depended on the dataset size.
\item  GAN-based models trained with larger sample sizes of balanced datasets exhibit less bias across various attributes. 
\end{itemize}

{\small
\bibliographystyle{ieee}
\bibliography{egbib}
}

\end{document}